\documentclass{article}

\usepackage[final]{corl_2019} 
\usepackage{graphicx}

\usepackage{booktabs}
\usepackage{xcolor,colortbl}
\usepackage{array}
\usepackage{subcaption}
\usepackage{amsmath}
\usepackage{multirow}
\usepackage{graphicx}
\usepackage{adjustbox}
\usepackage{amssymb}

\usepackage{hyperref}

\usepackage{mathtools}

\newcolumntype{L}[1]{>{\raggedright\let\newline\\\arraybackslash\hspace{0pt}}m{#1}}
\newcolumntype{C}[1]{>{\centering\let\newline\\\arraybackslash\hspace{0pt}}p{#1}}
\newcolumntype{R}[1]{>{\raggedleft\let\newline\\\arraybackslash\hspace{0pt}}m{#1}}

\title{Towards Learning to Detect and Predict Contact Events on Vision-based Tactile Sensors}
\author{
  Yazhan~Zhang \quad Weihao Yuan \quad Zicheng Kan \quad Michael Yu Wang\\
  Robotics Institute\\
  The Hong Kong University of Science and Technology \\
  Hong Kong, China\\
}

\begin{document}
\maketitle

\begin{abstract}

In essence, successful grasp boils down to correct responses to multiple contact events between fingertips and objects. In most scenarios, tactile sensing is adequate to distinguish contact events. Due to the nature of high dimensionality of tactile information, classifying spatiotemporal tactile signals using conventional model-based methods is difficult. In this work, we propose to predict and classify tactile signal using deep learning methods, seeking to enhance the adaptability of the robotic grasp system to external event changes that may lead to grasping failure. We develop a deep learning framework and collect 6650 tactile image sequences with a vision-based tactile sensor, and the neural network is integrated into a contact-event-based robotic grasping system. In grasping experiments, we achieved 52\% increase in terms of object lifting success rate with contact detection, significantly higher robustness under unexpected loads with slip prediction compared with open-loop grasps, demonstrating that integration of the proposed framework into robotic grasping system substantially improves picking success rate and capability to withstand external disturbances.

\end{abstract}

\section{Introduction}
\label{sec:introduction}

For human manipulation, neural receptors inside human fingers provide information of mechanical interaction and thus play a pivotal role in dexterous manipulations \cite{johansson1996sensory}. Similarly, artificial tactile sensors have been adopted and demonstrated to be effective in robotic systems for tasks including sensing object geometry \cite{johnson2009retrographic}, contact force \cite{vlack2005gelforce,yuan2015measurement,ma2018dense}, and detecting contact slippage \cite{yuan2015measurement,li2018slip}.
However, tactile sensing technology makes progress slowly for reasons including fabrication difficulties, limited resolution, multiplexing complexity, etc. \cite{dahiya2010tactile}. Apart from difficulties in hardware development, the inherent high dimensionality of tactile signals also challenges algorithms on information interpretation.

Among multi-modality tactile signals, detection of contact events (e.g. contact making, slippage, etc.) is irreplaceable for action adaptation. Also, anticipatory control policies support dexterous object manipulation by avoiding long time delays in human nervous system \cite{johansson1996sensory}. 
To mimic human touch feedback, multiple works have put efforts in integrating tactile sensing into contact events detection \cite{yuan2015measurement,dong2018maintaining}. However, few previous work has extensively studied contact event perception. Neither thorough contact event categorization nor generalizability of analytical methods is presented in previous works. Besides, for the nature of the high dimensionality of tactile signals, model-based methods have exceptional difficulties in interpreting useful contact information from raw readings. Data-driven methods are superior in learning patterns from high dimensional data. Therefore, some works have explored interpreting contact slip with learning methods \cite{li2018slip,veiga2015stabilizing,su2015force,van2018slip}. 
As far as we know, thorough contact event categorization and classification utilizing deep learning frameworks have not been explored, given its important role in reactive grasp manipulations.

Two major problems that hinder the processing effectiveness of tactile readings are: 1) unavailable suitable neural network; 2) lack of properly labelled tactile sequence data. 
Artificial functionality of sensing contact events that provides ground truth labelling for tactile sequences is not yet available as a tool. Therefore, we design a scheme taking advantage of the human sense of touch to label tactile sequences.
Towards the goal of tactile signal interpretation, we propose a neural network to process spatiotemporal readings and collect 6650 trials of contact sequence as a dataset. Raw tactile images are retrieved from FingerVision (first named in \cite{yamaguchi2016combining}) tactile sensor we developed \cite{zhang2018fingervision} (see Figure \ref{fig:training_objects}(a)) that outputs deformation images of the elastomer. A robot can employ the model in predicting and detecting contact events, preventing the robot from lifting target objects without enough contact and guaranteeing stable grasping by applying extra gripping force when unstable contact events appear. This work presents the first attempt to classify contact events into explicit 7 categories, and experimental results demonstrate that incorporating predictions and detections of contact events can substantially improve the grasping capability.


\textbf{Contributions}:
1) Collect a dataset containing vision-based tactile sequences with careful labelling by human expert is collected.
2) Propose a network capable of predicting and detecting tactile contact events that is critical to robotic grasping systems.
%


\begin{figure}
    \centering
    \includegraphics[width=0.98\textwidth]{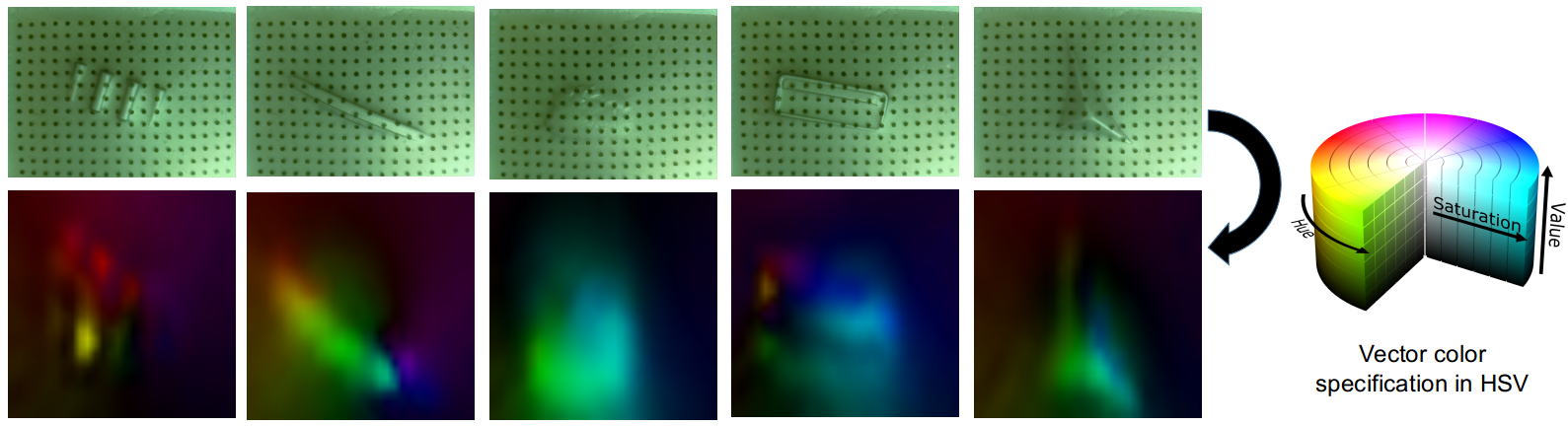}
    \caption{Examples of tactile raw images (first row) and the corresponding images of displacement vector fields (second row) collected by FingerVision. Displacement vectors are represented in HSV color specification space: Hue=Direction, Saturation=1, Value=Magnitude.}
    \label{fig:example_raw_disp}
    \vspace{-0.4cm}
\end{figure}

\section{Related Works}
\label{sec:related_works}

\textbf{Contact event detection.} Previous studies on human contact event detection have proven its importance for the interaction with environment. During a reaching and picking operation, four types of mechanoreceptors in human hand respond to contact events with different firing patterns, cooperatively extracting spatiotemporal features associated with mechanical contact events \cite{johansson1996sensory}. 
Most of the previous works on contact events studies focused on slip detection. Analytical methods with hand-crafted features have been presented in literatures.
Heyneman et al. \cite{heyneman2016slip} proposed two features based on spectral analyses extracted from dynamic tactile sensors that could be used to discriminate hand/object and object/world slip. 
Yuan et al. \cite{yuan2015measurement} presented entropy feature of the deformation fields from vision-based tactile sensor Gelsight to segment sensing area into slipping and stable regions with properly selected thresholds. 

Since the generalizability of these model-based methods with hand-crafted features 
were not tested in large number of repetitions and on different contact properties, data-driven methods, by comparison, are more appealing.
Su et al. \cite{su2015force} proposed to classify tactile signals into slip and stable categories with a lightweight multilayer perceptron (MLP) using BioTac tactile sensor \cite{wettels2008biomimetic}. However, classification performance was not adequate for robotic operations (accuracy around 80\%).
SVM \cite{cortes1995support}, random forest \cite{liaw2002classification}, Long-short-term-memory (LSTM) network \cite{hochreiter1997long} and convolutional LSTM (ConvLSTM) \cite{xingjian2015convolutional} have been applied to various tactile sensors \cite{veiga2015stabilizing, van2018slip, zapata2019learning} to generate slip/nonslip classifications.
In this work, we provide a finer categorization of contact events by borrowing insights of neuron firing patterns triggered by distingshed contact events during human manipulations \cite{johansson1996sensory}. Based on these categories, we propose a classification network and conduct extensive evaluations.

\textbf{Video prediction.} We are interested in predicting future frames of tactile image sequences in an unsupervised learning fashion. Currently, the state-of-the-art models of video prediction are PredNet \cite{lotter2016deep} with a inter-frame difference feed-forward operation, network in \cite{finn2016unsupervised} with pixel-transformation-based module called Dynamic Neural Advection (DNA), etc. ConvLSTM \cite{xingjian2015convolutional} units are building blocks of these two models, which extract spatial and temporal features simultaneously. While both two models successfully predict future frames with more natural looks and fewer defects, they are more suitable for videos involving only rigid body objects.
In our work, we propose a pixel motion network (``PixelMotionNet") that explicitly predicts the velocity of each individual pixel's value that is added afterwards to the current frame to obtain a future frame. This network shows superior performance on tactile sequence prediction by comparison.


\section{Learning Contact Event Detection and Predicting }
\begin{figure}
    \centering
    \includegraphics[width=0.9\textwidth]{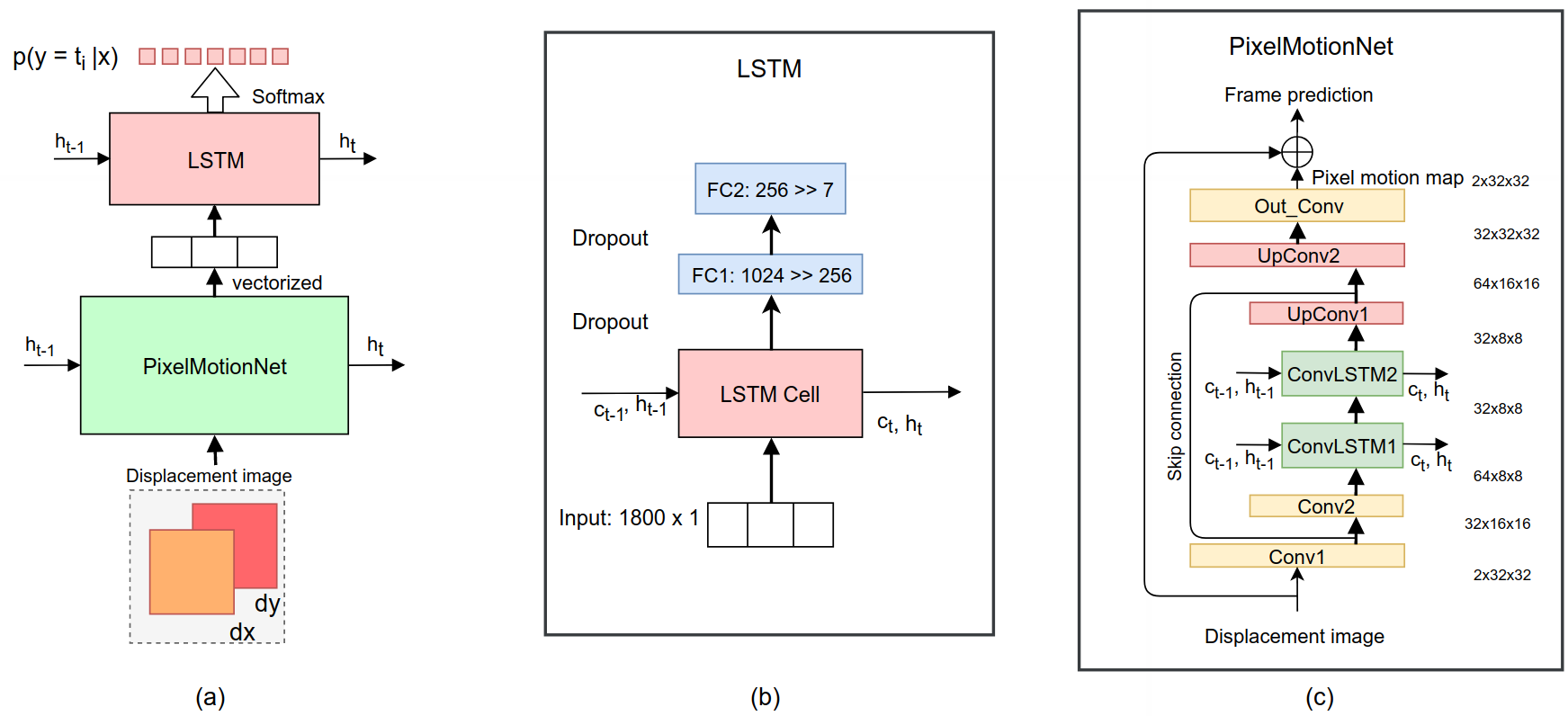}
    
    \caption{Network structure for contact events prediction. (a) Network structure.}
    \label{fig:network}
    \vspace{-0.3cm}
\end{figure}

In this work, we propose a LSTM-based neural network architecture as a spatiotemporal video sequence classifier for vision-based tactile data. In this section, a pipeline for tactile image sequence processing is introduced first and then description of the neural network structure is given.

\subsection{Tactile Image Preprocessing}
Instead of directly feeding the raw tactile images into neural networks \cite{calandra2017feeling}, we preprocess raw images (see Figure \ref{fig:example_raw_disp}) to acquire displacement fields. The sensor surface undergoes different deformations while making contact with different objects. 
By tracking the motions of the markers and then applying interpolation (for smoothness), the displacement vector fields are retrieved. In Figure \ref{fig:example_raw_disp}, 2-D displacement vector fields are color-coded in HSV space to facilitate better visualization, in which color and intensity represent direction and magnitude of vectors, respectively.


we extract tactile displacement images iteratively and then stack samples within a time window of $t_w$ as a sequence. Let $D[n]$ be the tactile image at $n^{th}$ sample in the sequence $S$, $N_s$ be the length of the tactile sequence and $f_s$ be the sampling frequency. $D[n]$ has two channels, $dX[n]$ and $dY[n]$, which are projection components of $V_d$ along X and Y axes respectively, on a grid of size $N_h \times N_w$. Then the collected sequence $S$ can be represented as

\begin{equation}
    S = \{D[n]\  |\  D[n]=\left(dX[n], dY[n]\right)^T, n=0, 1, ...,N_s\}
    \label{eq:sequence}
\end{equation}
where
\begin{equation*}
    \begin{split}
        dX[n]\{i, j\} &= \langle \vec{V}_d\{i, j\}, \vec{e}_1 \rangle \vec{e}_1 \\
        dY[n]\{i, j\} &= \langle \vec{V}_d\{i, j\}, \vec{e}_2 \rangle \vec{e}_2 \\
    \end{split}
\end{equation*}
$i=0,...,N_w$ and $j=0,...,N_h$ and $\vec{e}_1$, $\vec{e}_2$ denote the orthonormal bases for the X and Y axes on the sensor's coordinate system respectively.

In our configuration, the system runs at frequency $f_s=30$ Hz, and time window $t_w=1$ s, therefore each sequence contains 30 samples of deformation images. To further reduce the forward-propagation time of the network, we resample the original sequences with a stride of 2, thus the length of each resampled sequence $N_s=15$. After  interpolation, entries of each $dX[n]$ and $dY[n]$ are of shape $30\times30$. Succinctly, we have $D[n] \in \mathbb{R}^{C_h\times N_h \times N_w}$, where $C_h$ denotes the number of channels ($C_h=2$ in our case).




\subsection{Network Architecture}

The collected dataset is in the form of $I=\{S^1,...,S^K\}$, with $K$ being the number of samples. Corresponding to each sequence, label $y^i \in \mathbb{R}^C$ is in the one-hot encoding, where $C$ denotes the number of classes. The proposed network in Figure \ref{fig:network} consists of two subnetworks: Contact event detection network and video prediction network.

\textbf{Contact Event Detection Network.} The event detection network is a long-short-term-memory network (LSTM). Input to this network is sequences of tactile images $S_v = \{D_v[1], ...,D_v[N_s]\}$, where $D_v[n] \in \mathbb{R}^{M\times 1}$ and $M=1800$. 
To estimate the probabilities of a sequence $S_v$ that belongs to certain class, the last hidden state vector at position $n=N_s$ is fed into two cascaded Fully-Connected (FC) layers and a Softmax layer. 
To avoid overfitting in the training phase, two Dropout \cite{srivastava2014dropout} regularization layers with possibility of 0.5 are added, as depicted in Figure 2(b). Loss of the network is a reduced multi-class cross entropy between ground truth encodings and the corresponding predicted probability vectors.


%
%
%



\textbf{Video Prediction Network.} The network PixelMotionNet is composed of convolution/upsampling and ConvLSTM modules with a skip connection and an additive operation, as illustrated in Figure \ref{fig:network}(a). The model predicts the value expectation of each pixel in the next frame depending on spatiotemporal features propagated from previous image frames, inspired by the pixel transformation module presented in \cite{finn2016unsupervised, jaderberg2015spatial}. Upsampling layers recover feature maps back to the original size for the additive operation between the predicted pixel motion map and the current frame. Future frame prediction can be seen as a small modification by the pixel motion prediction map on the current frame. Different from \cite{finn2016unsupervised}, correlations between value changes of pixels in this study are only constrained by the sizes of the convolution kernels, instead of a spatial extent parameter. For multiple-frame prediction scenario, estimated next frames are circulated as new inputs back to the prediction network iteratively.

\begin{figure}
\centering
\begin{subfigure}[b]{0.25\columnwidth}
  \centering
  \includegraphics[height=3.0cm]{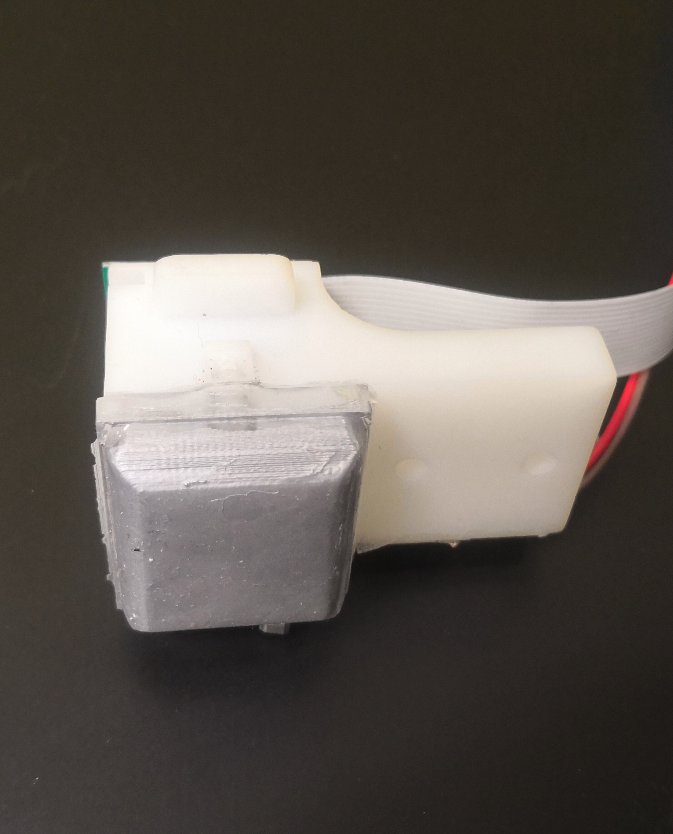}
  \caption{}
\end{subfigure}
\begin{subfigure}[b]{0.74\columnwidth}
  \centering
  \includegraphics[height=3.0cm]{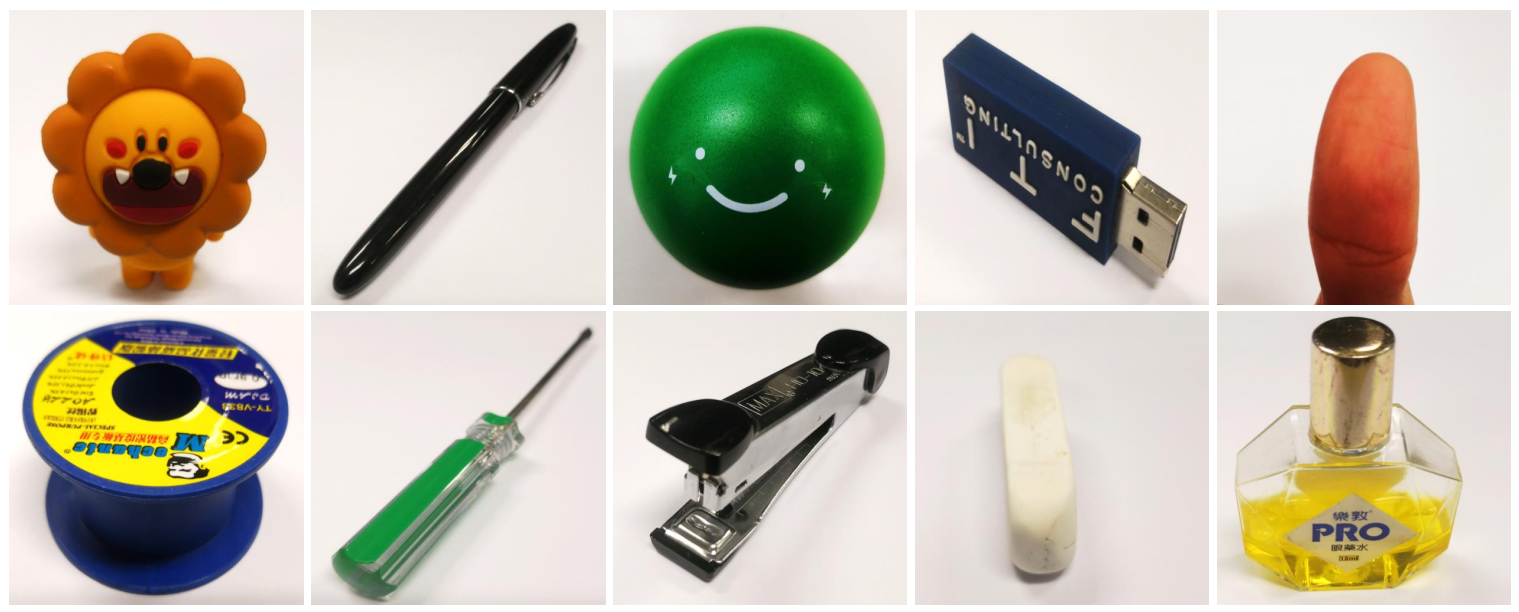}
  \caption{}
\end{subfigure}
    %
\caption{FingerVision sensor (a) and objects used for data collection (b).} 
\label{fig:training_objects}
\vspace{-0.5cm}
\end{figure}

The problem of spatiotemporal sequence prediction is to predict the most likely future sequence with length $N_{p}$ given the previous $N_{in}$ observations. For the PixelMotionNet, the loss function used during the training phase is $L^2$-norm of the difference between the ground truth images and the predicted ones. Suppose the future frame sequence is $\hat{S}$, $\hat{S}$ is computed by the following transformation function given an input sequence $S = \{D[1],...,D[N_{in}] \}$

\begin{equation}
    \hat{S} = \{\hat{D}[1+N_{in}],...,\hat{D}[N_{p}+N_{in}] \} = \mathcal{T}(S, \theta_v)
\end{equation}
and the loss is given by
\begin{equation}
    \mathcal{L} = \frac{1}{N_{p}}\sum^{N_{p}}_{k=1} (\hat{D}[k+N_{in}] - D[k+N_{in}])^2
\end{equation}
where $N_{in}$ and $N_{p}$ are lengths of the input frames and future frames, $\theta_v$ is the learned parameters of the prediction network, $\mathcal{T}$ is the transformation function of the PixelMotionNet.



\section{Data Collection}

Tactile image sequences span across temporal and spatial dimensions. When the tactile sensing area undergoes certain contact events, the tactile images evolve accordingly. In previous works, automatic and outcome-associated contact events labelling methods were adopted. In \cite{su2015force}, an IMU-based slip detecter was used to automatically annotate tactile readings during grasping into slip or nonslip. In \cite{li2018slip, veiga2015stabilizing, calandra2017feeling}, external cameras monitor relative motions between grippers and objects, or extra contact making/breaking detection networks are pre-trained as grasp success/failure discriminators. Object dropping from gripper is a consequence of unstable contacts, and the labelling of the tactile image with success or failure of grasping is reasonable only for stability estimation. However, dynamic grasping adjustment on the fly requires more temporal accuracy on labeling time windows during when contact events happen.
\begin{figure}
    \centering
    \includegraphics[width=0.8\textwidth]{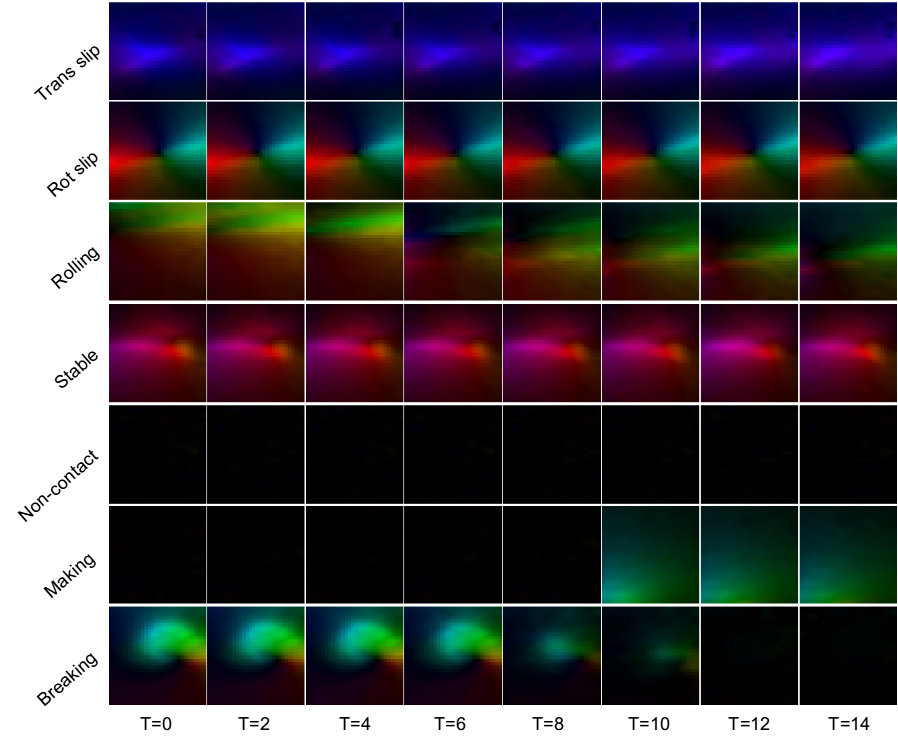}
    \caption{Example network input image sequences with different labels. Extra resampling with stride of 2 is applied for visualization and saving space.}
    \label{fig:network_input}
    \vspace{-0.3cm}
\end{figure}


In this work, labelling of data is handled by a human expert. The reason why human involved in  this procedure is that automatic data collection requires a  heuristic rule system or pretrained  discriminator to help label tactile squences, which is unavailable or asks for human prior to guide another labeling process to traine the discriminator beforehand. In comparison, labeling with the aid of human sense of touch is superior in temporal accuracy and more direct. When collecting each tactile sequence, one hand of the expert holds object and drives motion that associates to a targeted contact event on tactile sensing area. At the same time, the other hand triggers the corresponding labelling action for this sequence. According to \cite{johansson1996sensory}, for human, the time gap between tactile sensing of external stimuli and execution of the muscle contraction is roughly 100 ms, therefore, roughly 3 frames could be mislabelled with our annotation method. We classify contact events into 7 sets with unique contact behaviors spatiotemporally: 1) Translational slip: object sticks with the sensor surface and slips translationally; 2) Rotational slip: object sticks with the surface and slips rotationally; 3) Rolling: object rolls on the surface with contact maintained (for round edge) or experiences short contact breaking and remaking (for flat surface); 4) Stable: object moves with relatively small/no motion on the surface; 5) Noncontact: no object is making contact with the surface; 6) Making contact: object makes contact within the sequence; 7) Breaking contact: object breaks contact the current sequence.

To guarantee the generalizability of the network trained on the dataset to wide range of objects with different geometry, hardness, elasticity, texture, we selected 10 objects as shown in Figure~\ref{fig:training_objects}. Apart from properties of the selected objects, we applied forces and drove motions randomly by hand while collecting the data. Force intensity is in the range of 0$\sim$20Nand the duration is 0.5$\sim$2s for each interaction. For each class, we collected around 500 sequences. With skipping resampling (stride 2) and removal of out-of-class samples, we generated 6650 tactile sequences in total. Examples of tactile sequences with labels are shown in Figure~\ref{fig:network_input}. The collected dataset is publicly available at \href{url}{https://sites.google.com/view/tactile-event-corl2019/home}.

\section{Experiments}
\label{sec:exp}

In this section, we present settings of the network training and baseline comparisons first, then describe the implementation of reactive grasping experiments with integration of the proposed network.

\subsection{Network Training and Baseline Comparison}

We train separately the sub-networks considering better convergences and more convenient model evaluations. After separate training, our cascaded network takes in tactile image sequences and predicts jointly the probabilities of each class, as shown in Figure 2(a). All networks run on a computer with Intel i7-6700K CPU and NVIDIA GTX 1080Ti GPU, with batch size of 16, Adam optimizer \cite{kingma2014adam} and early stopping mechanism to prevent overfitting. To regularize the networks in the training phase, an extra weight decay with value of 0.05 is added. Considering the relatively small dataset size, small initial learning rates $4 \times 10^{-5}$ and $6 \times 10^{-5}$ are employed and decreased with training epoch for classification and video prediction network, respectively. For both training, dataset was split into training set and validation set with ratio of $9:1$ sampled randomly with a fixed random seed. After split, resulting support for each class is evenly spread.

For classification network, two additional baselines are selected. One is ConvLSTM, of which we flatten the output and then feed the vector to two FC layers. The another is a cascaded Convolution neural network (CNN) with a vanilla LSTM (CNN+LSTM). Three models are all built with relatively shallow structure bearing the goal to achieve real-time capability for decentralized processing of tactile units. 

For the video prediction network, quantitative evaluation of the proposed PixelMotionNet, state-of-the-art models PredNet \cite{lotter2016deep} and ConvLSTM \cite{xingjian2015convolutional} are presented. Models are trained only on a subset of 4 classes in the dataset without rolling, making contact, and breaking contact data, considering the degradation of video prediction performance when encountering these abrubtly varying events with hardly observable temporal coherences. Mean square error (MSE) and Structural Similarity Index (SSIM) \cite{wang2004image} are metrics used for evaluation.


\subsection{Performance Evaluation}
Following performance evaluations are all on the validation set. 

\textbf{Classification network.} Performances of LSTM and baseline models on tactile sequence classification are summarized in Table~\ref{tab:baseline_comparison}. Here evaluation metrics including forward propagation time $T_f$, accuracy, average precision, average recall, average F1-score over all classes, $N_{in}$ with which models achieved the best performances, and end epochs during training are given.

From the experimental results, LSTM outperforms the other two baseline models in all aspects except for the forward propagation time. ConvLSTM is superior in model size and acceleration on forward propagation for its weights sharing convolution layer compared to the fully connected structure in the vanilla LSTM. LSTM network overall achieves an accuracy peak of 98.50\% costing short forwarding time with $N_{in} = 12$.
\begin{table}[]
\centering
\small
\caption{Best performances and properties of models.}
\label{tab:baseline_comparison}
\setlength{\tabcolsep}{2.7mm}{
\begin{tabular}{cccccccc}
\toprule
Model & Acc(\%) & Prec (\%) & Rec (\%) & F1 (\%) & $T_f$ (ms) & $N_{in}$ & End Epoch \\ \midrule
ConvLSTM & 87.86 & 88.01 & 88.12 & 87.92 & \textbf{2.7} & 11 & 95\\
CNN+LSTM & 92.14 & 92.19 & 92.13 & 92.1 & 32.7 & 10 & 78\\ 
\textbf{LSTM} & \textbf{98.50} & \textbf{98.63} & \textbf{98.39} & \textbf{98.50} & 9.4 & 12 & 55\\ \bottomrule
\end{tabular}%
}
\vspace{-0.4cm}
\end{table}

\begin{table}[]
\centering
\small
\caption{Evaluation of models on predicting future frames, w.r.t. the future frame index.}
\label{tab:video_metrics}
\resizebox{\textwidth}{!}{%
\begin{tabular}{ccccccc}
\toprule
Model & Metrics & 1 & 2 & 3 & 4 & 5 \\ \midrule
\multirow{2}{*}{ConvLSTM} & MSE & 0.642$\pm$2.2930 & 0.940$\pm$3.600 & 1.310$\pm$5.7200 & 1.560$\pm$7.4700 & 1.800$\pm$9.4300 \\
 & SSIM & 0.950$\pm$0.0100 & 0.940$\pm$0.0200 & 0.970$\pm$0.0300 & 0.850$\pm$0.0400 & 0.830$\pm$0.0400 \\
\multirow{2}{*}{PredNet} & MSE & 0.304$\pm$0.1580 & 0.259$\pm$0.1660 & 0.341$\pm$0.2110 & 0.483$\pm$1.0900 & 0.649$\pm$1.3900 \\
 & SSIM & 0.976$\pm$0.0009 & 0.974$\pm$0.0040 & 0.967$\pm$0.0040 & 0.956$\pm$0.0059 & 0.940$\pm$0.0069 \\ 
\multirow{2}{*}{\textbf{PixelMotionNet}} & MSE & \textbf{0.024$\pm$0.0005} & \textbf{0.061$\pm$0.0049} & \textbf{0.107$\pm$0.0142} & \textbf{0.179$\pm$0.1060} & \textbf{0.301$\pm$0.1680} \\
 & SSIM & \textbf{0.990$\pm$0.0001} & \textbf{0.988$\pm$0.0002} & \textbf{0.979$\pm$0.0006} & \textbf{0.970$\pm$0.0018} & \textbf{0.957$\pm$0.0029} \\ \bottomrule
\end{tabular}%
}
\vspace{-0.4cm}
\end{table}

\begin{figure}
    \centering
    \includegraphics[width=0.9\textwidth]{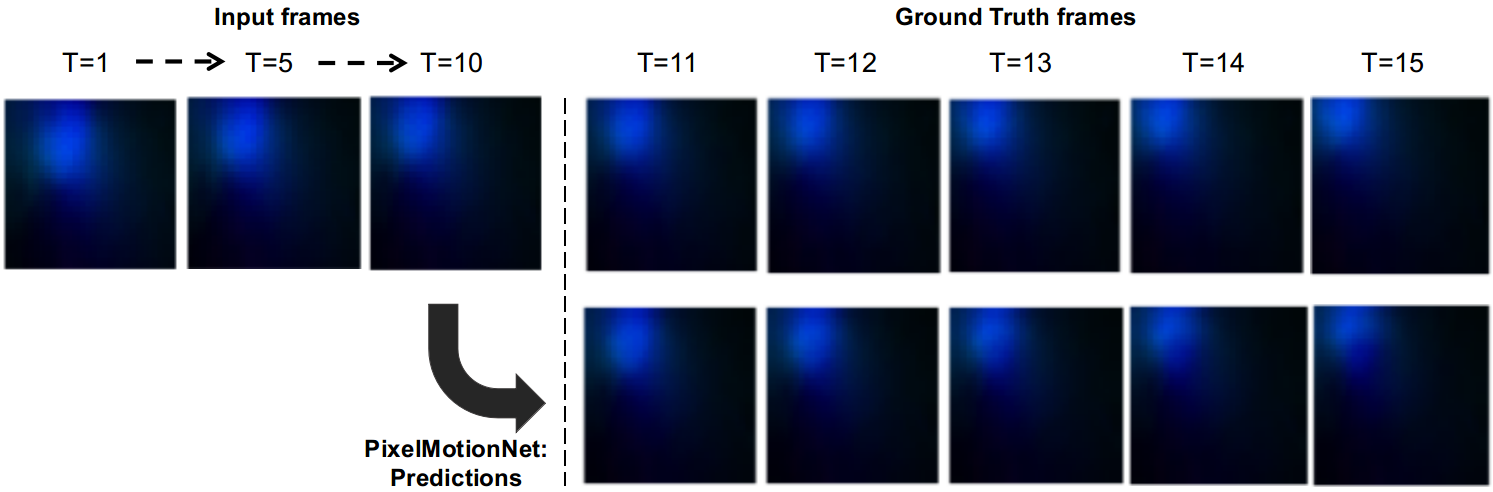}
    \caption{Tactile sequence prediction by PixelMotionNet: ground truth sequence vs. predicted sequence.}
    \label{fig:image_prediction}
    \vspace{-0.4cm}
\end{figure}

\textbf{Video prediction network.} Quantitatively, from the results in Table~\ref{tab:video_metrics}, PixelMotionNet is superior in both metrics compared to the other two baselines. We notice that the farther the network predicts, the larger the divergence between the prediction and the ground truth is. Furthermore, variations of predictions rise as future frame index increases, which is reasonable since the pixel value expectations of the future frames are less predictable as the network predicts further into the furture. Qualitatively, in Figure~\ref{fig:image_prediction}, the predicted future frames are illustrated aside with the ground truth frames to show how well the prediction network performs on the validation set. The PixelMotionNet captures the motion of force concentration and pixel values successfully (the result is more than copying the last image of the input sequence). It can be also seen that as the models predict more frames into the future, blurrier the images become, which is consistent with the quantitative analysis.

\subsection{Experiments in Real-time Grasping}

\begin{figure}[thb]
\centering
\begin{subfigure}[b]{0.4\columnwidth}
  \centering
  \includegraphics[height=3.5cm]{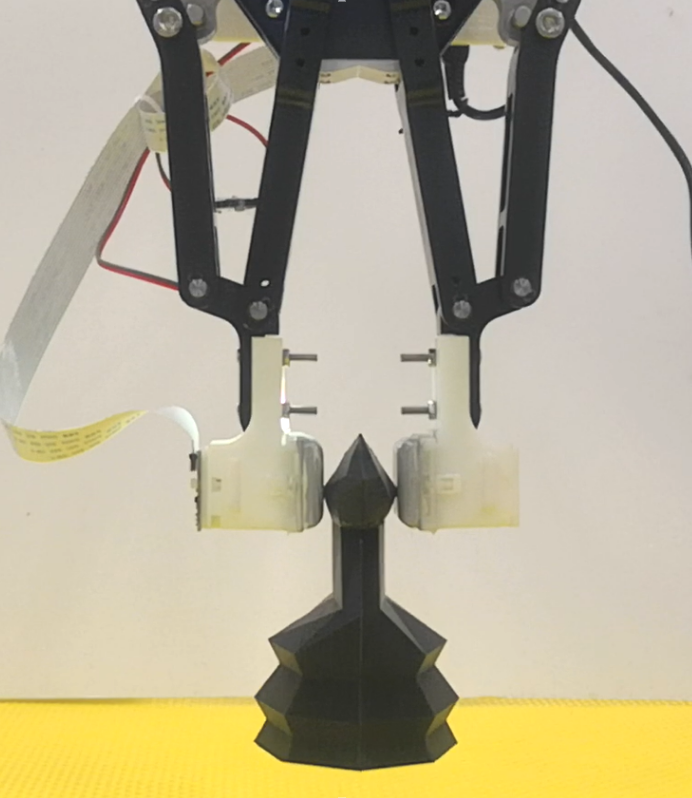}
  \caption{Experiment scene}
\end{subfigure}
\begin{subfigure}[b]{0.4\columnwidth}
  \centering
  \includegraphics[height=3.5cm]{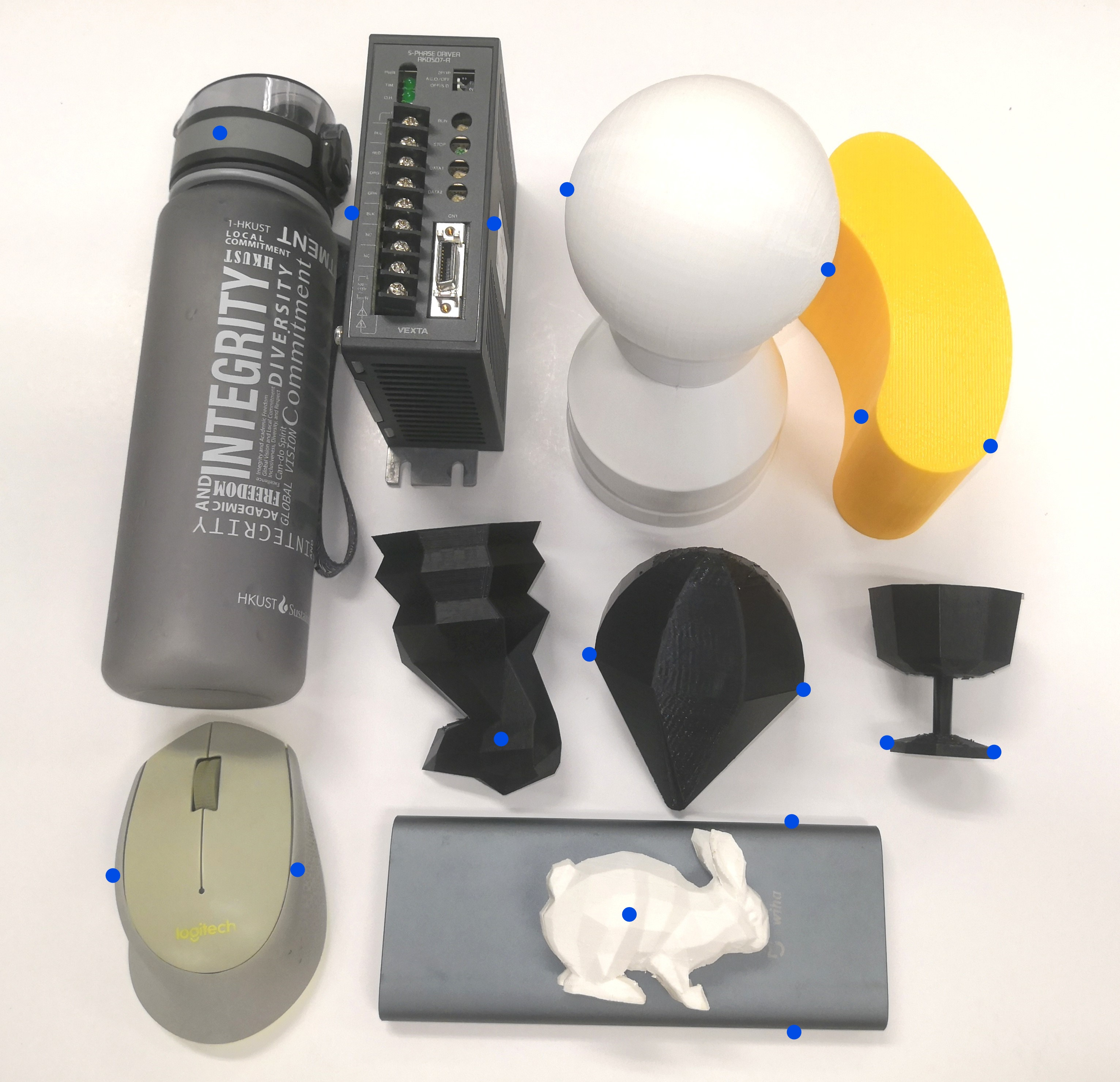}
  \caption{Grasping objects}
\end{subfigure}
\caption{Real-time grasping experiments are performed on 10 distinct-shape objects. Grasping sites are denoted by blue dots in (b).}
\label{fig:grasp_scene_obj}
\vspace{-0.4cm}
\end{figure}

Since what we concern more in practice is how well the contact event prediction and detection networks help robotic manipulation, we directly perform real-time grasping experiments integrating the proposed network instead of evaluating it on a test dataset. In the grasping experiments, we install the FingerVision sensor on a Robotiq 2-finger gripper mounted on a UR10 robotic arm, as shown in Figure~\ref{fig:grasp_scene_obj}(a). This section is supplemented with a video document.

\textbf{Contact detection experiment.} To evaluate the performance of our framework in detecting contact making, we test the grasping success rate with and without our tactile contact detection on 10 objects with different shapes, sizes, and materials, as presented in Figure \ref{fig:grasp_scene_obj}(b). We first choose the grasping sites and measure the required gripper openings with ruler manuly, with which the objects can be narrowly grasped and lifted. Then we add a small noise with standard deviation $\sigma_n=1$ mm to simulate noises in non-contact measurements, e.g., vision, and test if the gripper can grasp and lift the objects. For open-loop operation, the gripper closes until it reaches target opening, while for close-loop grasp, the gripper adjusts its opening until the network indicates contact making event. 10 trials are executed on each object in both groups. The number of successful trials for each object and average success rate are summarized in Table~\ref{tab:grasping}. The success rate of these 10 objects is $46\%$ without contact detection while $98\%$ with the contact detection, reflecting that the contact detection facilitates the grasping significantly.

\begin{figure}[thb]
\centering
\begin{subfigure}[b]{0.25\columnwidth}
  \centering
  \includegraphics[height=2.5cm]{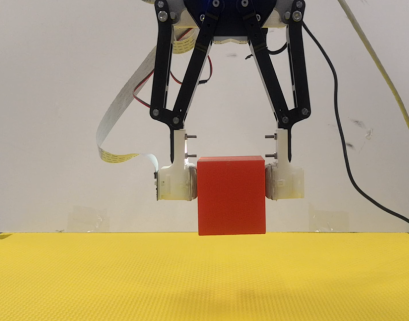}
  \caption{Initial state}
\end{subfigure}
\begin{subfigure}[b]{0.25\columnwidth}
  \centering
  \includegraphics[height=2.5cm]{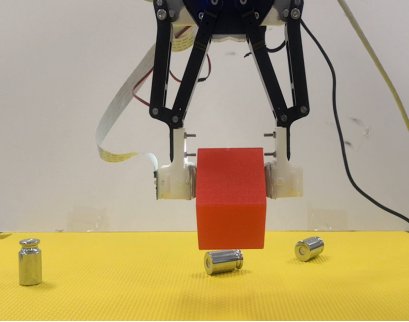}
  \caption{Slip occurs}
\end{subfigure}
\begin{subfigure}[b]{0.25\columnwidth}
  \centering
  \includegraphics[height=2.5cm]{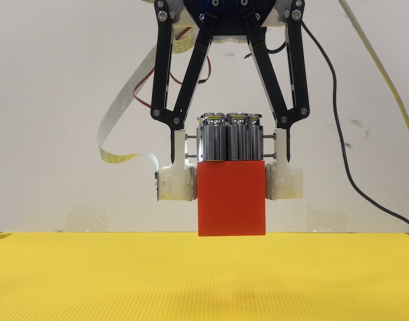}
  \caption{Stable grasp}
\end{subfigure}
\caption{Slip happens in (b) without slip prediction while grasp remains stable with slip prediction in (c) under increasing load. (Better shown in video)}
\label{fig:slip_sustain}
\vspace{-0.4cm}
\end{figure}

\textbf{Stable grasp under slippage.} Complementary to the above experiments, a further ability test of the proposed framework to help stabilize grasped object under external disturbance by predicting and detecting slip occurrence is performed. In this experiment, a gripper with the tactile sensor holds the object, then weights are loaded on top of the object one by one to trigger slip on the contact surfaces, as illustrated in Figure~\ref{fig:slip_sustain}. Different outcomes under increasing loads are given in Figure \ref{fig:slip_sustain}(b) and (c), without and with slip prediction by our framework respectively. With same initial gripper opening, while gripper lost stable contact after 3 weights were loaded without slip prediction, it maintained stable grasp with all 7 weights loaded by actively controlling the gripper's opening when slip prediction is provided. Qualitatively, slip prediction mechanism enhances the ability of the grasping system under external disturbances.

Overall, the experimental results suggest that the proposed framework is able to cover multiple phases of robotic grasping and enhance grasping system performance substantially.

\setlength{\tabcolsep}{4pt}
\begin{table}[t]
\centering
\small
\caption{Grasping Success Rates}
\begin{tabular}{c c c c c c c c c c c c c}
\toprule
Detection & Box & Ball & Chess & Yellow & Bottle & Cup & Cone & Rabbit & Mouse & Power & Average \\
\midrule
Without & $2$ & $3$ & $6$ & $6$ & $2$ & $6$ & $5$ & $6$ & $5$ & $5$ & $46\%$ \\
\rowcolor[gray]{1.0}
With & $10$ & $10$ & $10$ & $9$ & $10$ & $10$ & $10$ & $10$ & $9$ & $10$ & $\mathbf{98\%}$ \\
\bottomrule
\end{tabular}
\label{tab:grasping}
\vspace{-0.5cm}
\end{table}
\section{Conclusion and Discussion}
\label{sec:conclusion}

Humans manipulate objects with smooth transitions between contact phases by predicting and detecting contact events. In this paper, we try to endow the robot with similar capabilities. To this goal, we develop a contact event prediction and detection network, consisting of classification and sequence prediction subnetworks. We collect a contact sequence dataset of size 6650 with careful labelling by a human expert. Taking a separate training and evaluation scheme, the results show that the subnetworks outperform baselines on inference tasks given tactile sequences. Jointly, we integrate the networks together and implement real grasp experiments, of which the results show that the proposed framework grant robotic grasping system with new skills and improve overall performance.

In our work, we attempt to predict and detect contact events in the absence of visual modality and proprioceptive input. However, we observe that for contact making and breaking, PixelMotionNet is incapable of capturing changes that happen in relatively short time, which leads to an inaccurate prediction. On one hand, this stems from limited frame rate of hardware set up. On the other hand, vision and proprioceptive signals could potentially alleviate this problem by building a robotic perception system based on multi-modality sensor fusion. This raises an interesting direction as our future work.



\acknowledgments{This work is supported by the Hong Kong Innovation and Technology Fund (ITF) ITS-018-17FP.}

\clearpage
\bibliography{example}  

\end{document}